\documentclass[a4paper, 10 pt, conference]{ieeeconf}  
\IEEEoverridecommandlockouts                              

\usepackage{graphicx} 
\usepackage{mathptmx} 
\usepackage{times} 
\usepackage{amsmath} 
\usepackage{amssymb}  
\usepackage{multirow}
\usepackage{url}
\usepackage{bm}
\usepackage{booktabs}
\usepackage{cite}
\usepackage{pifont}
\usepackage{stfloats}

\newcommand{\cmark}{\ding{51}}

\begin{document}

\title{\LARGE \bf
MR-GLi: Mixed Reality-Based Gripper-Linked Overlays for Underwater Robot Arm Teleoperation via Bilateral Control
}

\author{Masashi Sasago$^{1\dag}$, Masato Kobayashi$^{1,2,3\dag*}$, Yuki Uranishi$^{1,2}$ 
\thanks{
${\dag}$ Equal Contribution,
$^{1}$ Graduate School of Information Science and Technology, The University of Osaka, $^{2}$ D3 Center, The University of Osaka, $^{3}$ Graduate School of Maritime Sciences, Kobe University, * corresponding author: kobayashi.masato.cmc@osaka-u.ac.jp This work involved human subjects. Approval of all ethical and experimental procedures and protocols was granted by D3 Center, The University of Osaka, under Application No. 2024-11.}
}

\maketitle
\begin{abstract}
Visual torque feedback supports underwater bilateral teleoperation, but the benefit of
mixed reality (MR) over conventional monitor presentation remains unclear. We present
MR-GLi, an MR interface that spatially registers a reaction torque indicator and
wrist-camera image to the robot gripper. Twenty participants performed lift and
pick-and-place tasks with rigid and compliant objects in a counterbalanced within-subject
comparison with a 2D monitor, using identical visual-feedback content and four-channel
bilateral control. MR-GLi provided gripper-linked access to visual feedback while maintaining a similar level of torque-regulation performance to the 2D monitor.
Subjective
evaluation further indicated reduced perceived burden associated with shifting attention
between the workspace and visual feedback. These results demonstrate the feasibility of
gripper-linked MR overlays for underwater bilateral teleoperation and highlight the
importance of considering information access in addition to task performance.
Additional material: \url{https://mertcookimg.github.io/mr-gli/}
\end{abstract}

\section{INTRODUCTION}
Underwater manipulators enable intervention in environments where direct human access
is risky and costly~\cite{brown2025rov,sitler2024uvms,liu2025aqua}. Bilateral control conveys
environmental reaction forces to the operator~\cite{mk2025alpha}, but underwater
disturbances such as drag and buoyancy can make the small torques required for grasp
regulation difficult to perceive~\cite{tsunoori2026biaquabilateralcontrolbasedimitation}. Visual force feedback can complement haptic
perception~\cite{moortgat2022rift,mielke2025sensary}. However, when such information is
presented on a separate display, operators may need to shift their visual attention
between the feedback and the workspace. How visual feedback is made accessible during
manipulation is therefore an important consideration in teleoperation interface design.

\begin{figure}[t]
    \centering
    \includegraphics[width=\linewidth]{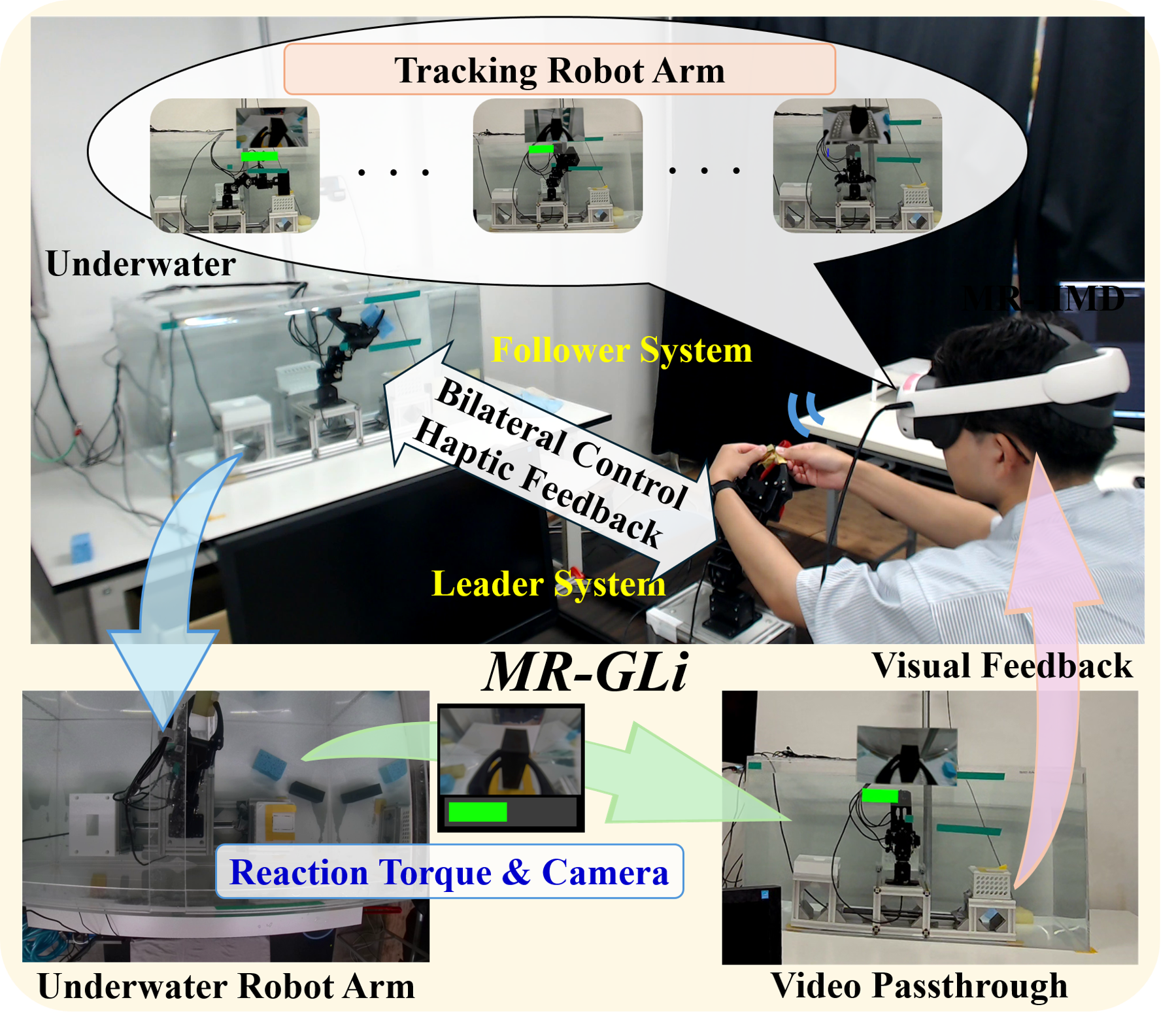}
    \caption{MR-GLi in operation. Inside the head-mounted display, the reaction
    torque indicator and wrist-camera image are registered to the gripper and
    follow the arm as it moves (inset). The monitor condition presents the same
    visual information on a display beside the tank.}
    \label{fig:teaser}
\end{figure}

Mixed reality (MR) provides a means of presenting such visual feedback within the
operator's field of view. MR-UBi~\cite{mrubi} previously evaluated a reaction torque
indicator (RTI) under MR-HMD video passthrough. Both experimental conditions used the
same bilateral controller and passthrough view, while the RTI was presented only in
the proposed condition. The results showed that adding the RTI improved underwater
torque regulation. However, the RTI was displayed at a fixed position in the HMD
view, and no wrist-camera image was provided. Thus, although MR-UBi demonstrated the
benefit of visual torque feedback, how such information should be spatially presented
during manipulation and how an MR-based interface compares with conventional monitor
presentation remained open questions.

Building on MR-UBi, we propose \emph{MR-GLi} (Mixed Reality-Based Gripper-Linked Overlays for Underwater Robot Arm Teleoperation via Bilateral Control), an MR interface that adds a wrist-camera view and spatially registers both
the camera image and the RTI to the robot gripper (Fig.~\ref{fig:teaser}). The two
visual elements therefore move with the manipulation site rather than remaining fixed
in the operator's view. We evaluate MR-GLi against a conventional 2D monitor positioned
beside the workspace while keeping the four-channel bilateral controller, RTI encoding,
and wrist-camera content identical. In the monitor condition, the operator directly
observes the workspace and consults the external display; in MR-GLi, the workspace is
viewed through video passthrough and both visual elements remain spatially associated
with the gripper. Accordingly, the comparison evaluates the two complete display
configurations rather than isolating overlay placement alone.

Prior research suggests that spatial proximity between related information can reduce
the effort required to access and integrate that information, although it does not
necessarily improve task accuracy~\cite{wickens1995pcp,warden2023overlay}. Gripper-linked
presentation may therefore reduce the need for visual attention shifts even when
torque-regulation performance is comparable to that achieved with a monitor. At the
same time, an MR-HMD introduces potential disadvantages, including headset weight,
restricted field of view, and video-passthrough degradation. Evaluating both objective
task performance and operator experience is therefore necessary to characterize the
practical trade-off between the two display configurations.

We investigate this trade-off in a counterbalanced within-subject study with twenty
participants. Participants performed lift and pick-and-place tasks with rigid and
compliant objects under both display configurations, yielding 160 trials in total.
We evaluate torque regulation and task time as objective measures, together with
perceived gaze-shift burden, workload, and usability as subjective measures.

The contributions of this paper are:
\begin{itemize}
    \item \emph{MR-GLi}, a gripper-linked MR interface that extends underwater
    bilateral teleoperation with spatially registered reaction-torque feedback and
    wrist-camera imagery.
    \item A counterbalanced within-subject comparison between MR-GLi and a conventional
    2D monitor, with bilateral control and visual-feedback content held constant.
    \item Experimental evidence that MR-GLi significantly reduces perceived gaze-shift
    burden, while no significant differences are detected in torque-regulation
    performance, workload, or overall usability.
\end{itemize}

\section{RELATED WORK}
\begin{figure*}[t]
    \centering
    \includegraphics[width=\linewidth]{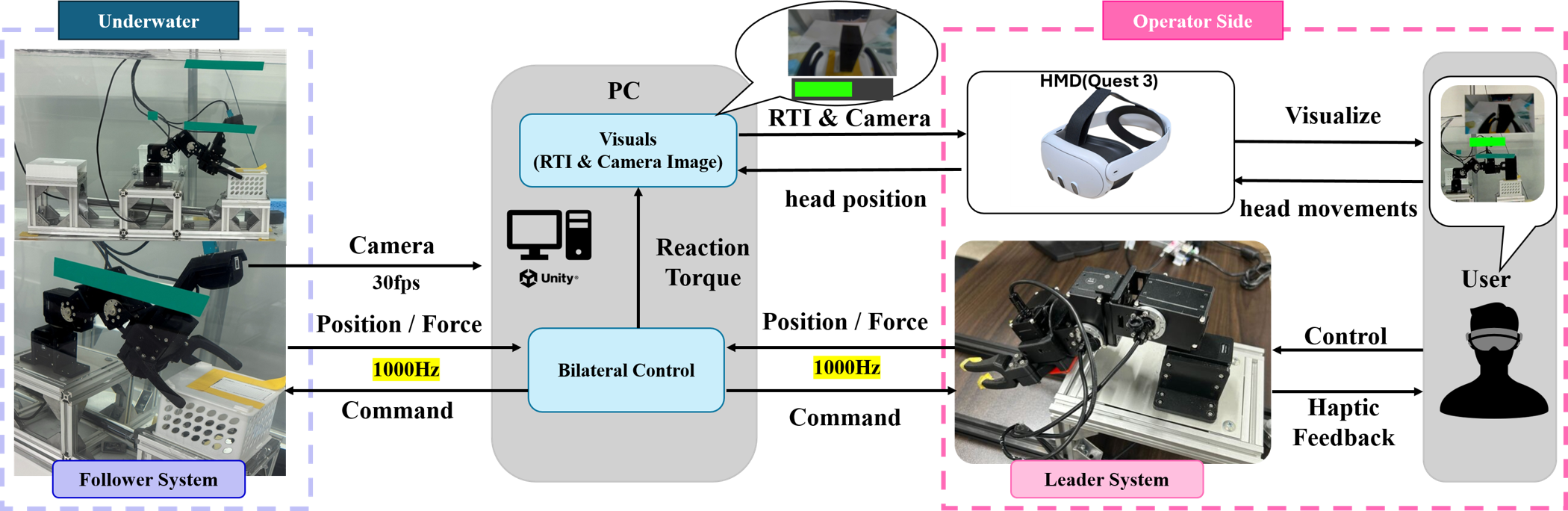}
    \caption{System overview of MR-GLi and the 2D Monitor configuration.}
    \label{fig:system}
\end{figure*}
\subsection{Visual Force Feedback in Telemanipulation}
Visual force feedback supports telemanipulation when haptic feedback is unavailable
or difficult to interpret. A-RIFT~\cite{moortgat2022rift} represents interaction force
using a bar overlaid on the camera image, providing sensory substitution through a
conventional computer interface. SensARy~\cite{mielke2025sensary} investigates AR-based
force visualization for touchless robot control, including alternative visualization
placements. These studies establish visual feedback as a means of conveying contact
information without direct haptic feedback, but do not address its presentation during
underwater bilateral teleoperation.

Underwater systems such as Ocean One~\cite{oceanone2016khatib} and disturbance-observer-based
bilateral controllers~\cite{motoi2023twobiuwr} provide haptic feedback during remote
manipulation. In this setting, visual torque feedback can complement the haptic channel
to support grasp regulation. MR-UBi~\cite{mrubi} demonstrated improved torque regulation
by adding an RTI to an underwater bilateral teleoperation interface. However, its
comparison used the same HMD in both conditions and therefore evaluated the benefit of
the RTI, leaving the relative value of MR and conventional monitor presentation
unresolved. MR-GLi addresses this question by comparing the two display configurations
with the bilateral controller, RTI, and wrist-camera image held constant.

\subsection{Spatial Presentation and Visual Information Access}\label{sec:place}
The proximity compatibility principle~\cite{wickens1995pcp} suggests that spatial
proximity supports information integration, whereas separation can benefit focused
attention. Warden et al.~\cite{warden2023overlay} found that overlaid displays improved
response time without compromising accuracy in integration tasks, while separate
displays benefited accuracy in focused-attention tasks. Head-up display studies also
report poorer detection of unexpected events~\cite{fadden1998hud}, and AR-HMD research
links spatial separation to information-access costs~\cite{poole2026iac}. Together,
these findings indicate that the benefits of display proximity depend on task demands.

Underwater grasp regulation requires integrating the gripper state, visual torque
feedback, and haptic feedback. Both configurations in this study present the RTI and
wrist-camera image together; MR-GLi additionally registers them to the gripper.
Prior work motivates testing whether this configuration reduces perceived visual-access
burden while preserving torque regulation, but does not establish this outcome for
underwater bilateral control. We therefore evaluate both dimensions through a comparison of complete display configurations.

\section{Method}

\subsection{System Overview}
MR-GLi is built on the MR-UBi platform~\cite{mrubi}: a low-cost 3-DoF underwater arm
with a two-fingered gripper, four-channel leader--follower bilateral control providing
haptic feedback, and the RTI. Two things are new here: a camera at the follower wrist, and the registration of
both visual elements to the gripper frame so that they travel with it. MR-UBi rendered its RTI at a fixed position in the head-mounted field of view
and had no wrist-camera image. Fig.~\ref{fig:system} shows how the parts fit together.

\subsection{Robot Platform and Bilateral Control}
\begin{figure}[t]
    \centering
    \includegraphics[width=\linewidth]{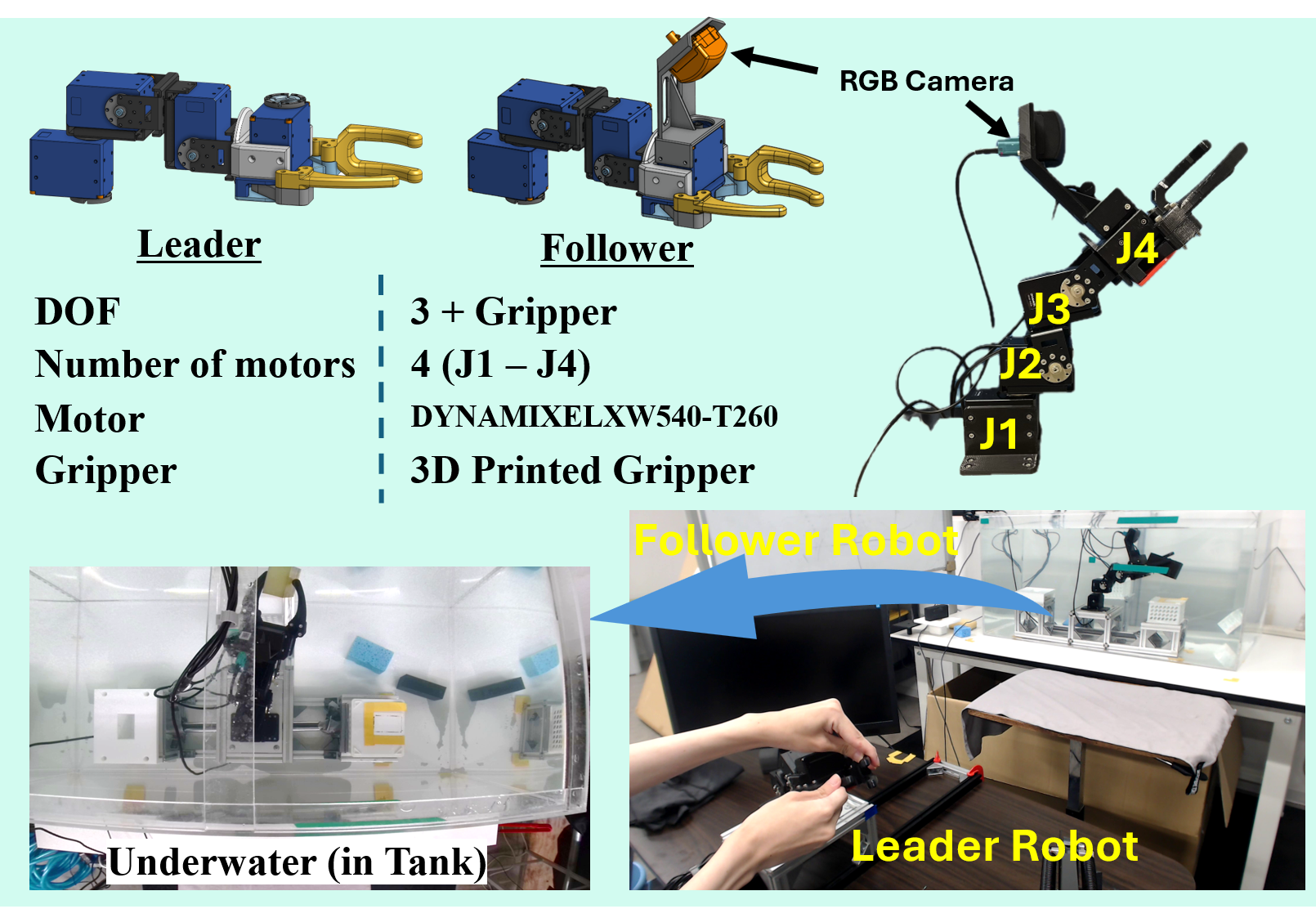}
    \caption{Leader and follower arms, sharing three rotational joints (J1--J3)
and a two-fingered gripper (J4) driven by waterproof Dynamixel XW540-T260 actuators. An
RGB camera sits at the follower wrist; only the follower is submerged.}
    \label{fig:arm}
\end{figure}

Each joint uses a fully waterproof actuator (Dynamixel XW540-T260, IP68). The
mechanism has three rotational joints and a two-fingered gripper, inspired by the
structure of~\cite{yamamoto2024gripper}; links and frames are aluminum profiles and
3D-printed parts (Fig.~\ref{fig:arm}).

Bilateral control is realized by enforcing position tracking and the action--reaction
law,
\begin{align}
\theta_l - \theta_f &= 0, \label{eq:pos}\\
\tau_l + \tau_f &= 0, \label{eq:force}
\end{align}
where $\theta$ and $\tau$ denote joint angle and torque, and subscripts $l$ and $f$
denote leader and follower. Joint torque is estimated sensorlessly with a reaction
torque observer (RTOB)~\cite{RTOB}. The control loop is approximately
$1\,\mathrm{kHz}$ on all four
joints. The displayed quantity is the gripper-joint reaction torque, i.e.\ the same
signal used inside the bilateral loop; we refer to it as the grasping torque.

\subsection{Visual Feedback Components}
\begin{figure}[t]
\centering
\includegraphics[width=\linewidth]{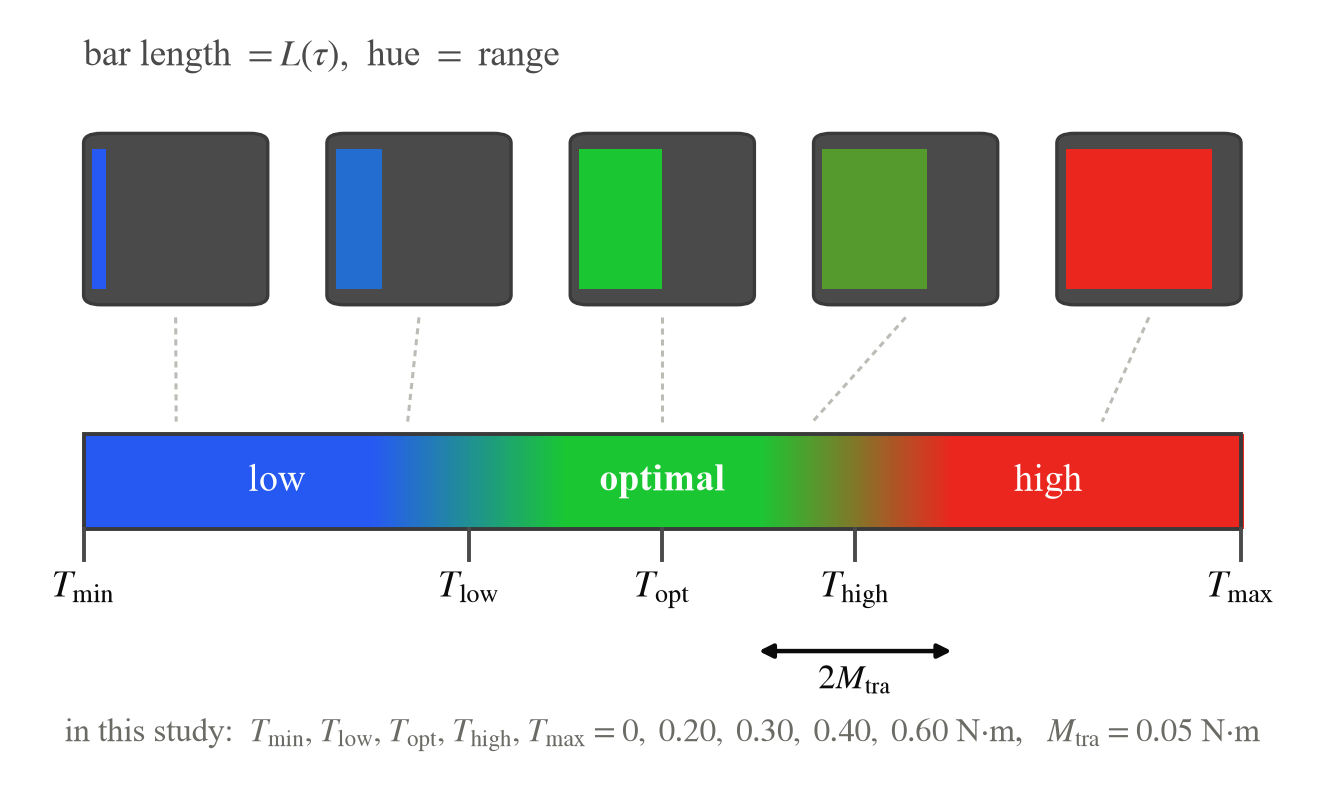}
\caption{The RTI encoding, shared by both configurations. Bar length is the
filling ratio of Eq.~\eqref{eq:fill}; hue follows Eq.~\eqref{eq:hue}.}
\label{fig:rti}
\end{figure}

Two elements are shown to the operator. The wrist camera is an RGB camera mounted on
the follower arm just behind the gripper, giving a close view of the gripper and the
object it holds. The RTI encodes torque with two cues: bar length for magnitude and hue for state.
The filling ratio is
\begin{equation}
L(\tau) = \frac{\min(\max(\tau, T_{\min}), T_{\max}) - T_{\min}}{T_{\max} - T_{\min}}
\times 100 ,
\label{eq:fill}
\end{equation}
and hue reports which range the torque is in,
\begin{equation}
c(\tau)=
\begin{cases}
c_{\mathrm{low}}, & \tau < T_{\mathrm{low}},\\
c_{\mathrm{opt}}, & T_{\mathrm{low}} \le \tau \le T_{\mathrm{high}},\\
c_{\mathrm{high}}, & \tau > T_{\mathrm{high}},
\end{cases}
\label{eq:hue}
\end{equation}
with each threshold crossed by a linear blend of width $2M_{\mathrm{tra}}$ so
that gradual torque changes remain perceptible while the optimal band stays
categorically distinguishable~\cite{mrubi}. Parameters are
$T_{\min}=0.00$, $T_{\max}=0.60$, $T_{\mathrm{opt}}=0.30\,\mathrm{N{\cdot}m}$, optimal band
$[T_{\mathrm{low}},T_{\mathrm{high}}]=[0.20,0.40]\,\mathrm{N{\cdot}m}$,
$M_{\mathrm{tra}}=0.05\,\mathrm{N{\cdot}m}$, with blue, green, and red for the three
ranges (Fig.~\ref{fig:rti}). These values are identical in both conditions and, as
in~\cite{mrubi}, are applied unchanged to both objects: the band is a common reference
target, not a per-object slippage or overload threshold, so the torque measures below
report regulation to a specified band rather than validated grasp quality.

\begin{figure}[t]
\centering
\includegraphics[width=0.9\linewidth]{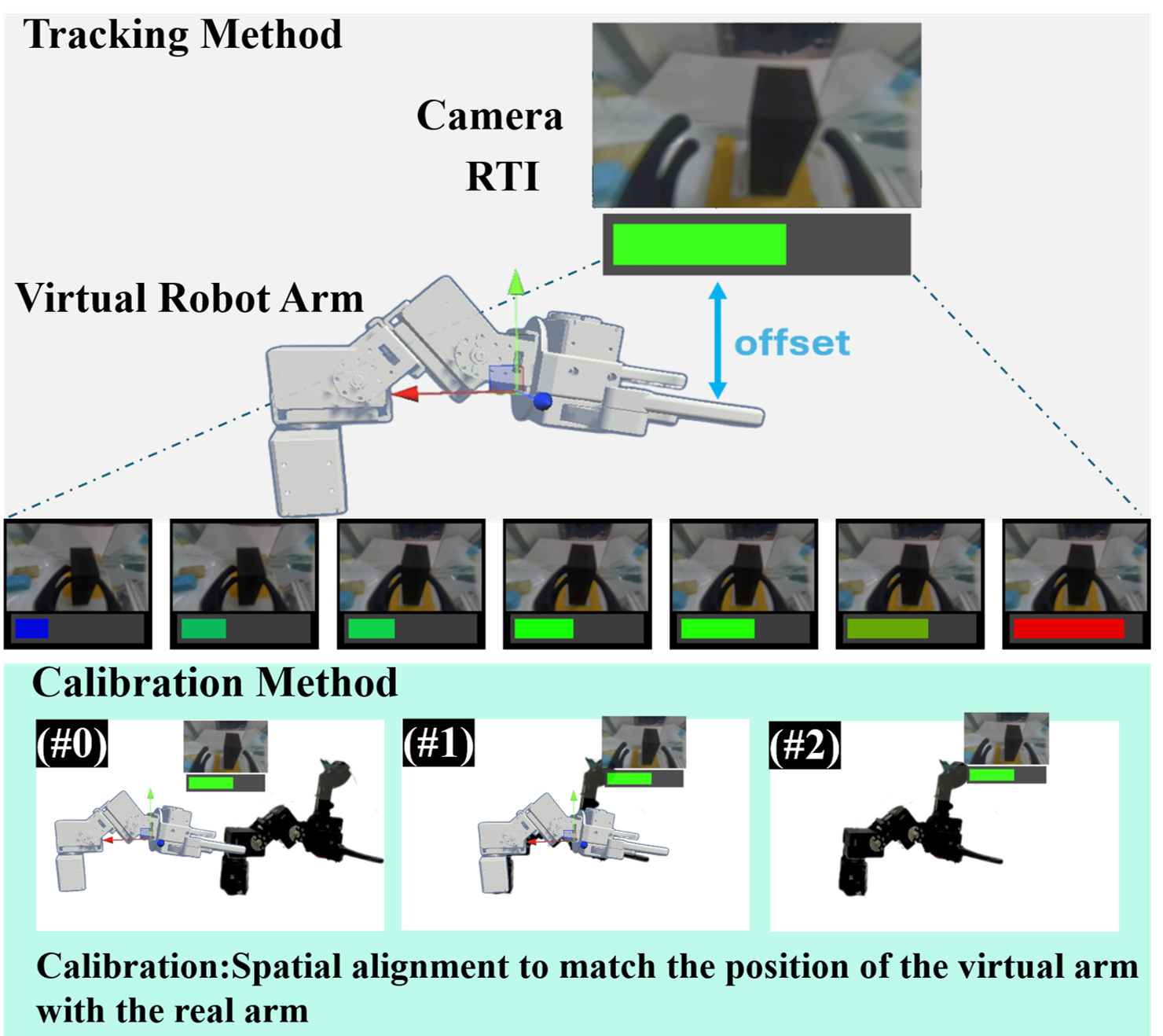}
\caption{Tracking and calibration in MR-GLi.}
\label{fig:mrglo}
\end{figure}
\subsection{Gripper-Linked Overlays and Calibration}\label{sec:glo}

The poses of both visual panels are determined from the follower arm's joint
encoder readings (Fig.~\ref{fig:mrglo}, top). Forward kinematics computes the
gripper pose $T_g(\mathbf{q})$ from the joint angles $\mathbf{q}$. The wrist-camera
image and RTI are then positioned relative to the gripper frame at a fixed offset
$\mathbf{t}_o$,
\begin{equation}
\mathbf{p}(\mathbf{q}) = T_r\,T_g(\mathbf{q})\,\mathbf{t}_o,
\label{eq:overlay}
\end{equation}
where $T_r$ represents the rigid transformation that registers the robot coordinate
frame to the MR coordinate frame. Because the panel pose depends on
$T_g(\mathbf{q})$, the visual information follows the gripper as the arm moves.
The wrist-camera image and RTI are arranged on a single world-space canvas at a
fixed offset above the gripper. The canvas is oriented toward the operator at each
frame so that the visual information remains legible regardless of the gripper
orientation.

The transformation $T_r$ is determined once before each experimental session by
manual spatial alignment (Fig.~\ref{fig:mrglo}, bottom). First, the physical robot
arm and a virtual model of the robot arm are displayed simultaneously
(Fig.~\ref{fig:mrglo}, \#0). The user then adjusts the virtual model by
translation along the three axes and yaw rotation about the vertical axis until it
is spatially aligned with the physical arm (\#1). After the alignment is confirmed,
the transformation is stored and the virtual model is hidden (\#2). The position is
stored as an offset from the centroid of the headset's play-area boundary, while the
orientation is stored as a quaternion. Because the play-area boundary is fixed to
the room rather than to the wearer, the registered coordinate frame can be restored
when the application is launched without repeating the alignment procedure.

\begin{figure*}[t]
    \centering
    \includegraphics[width=\linewidth]{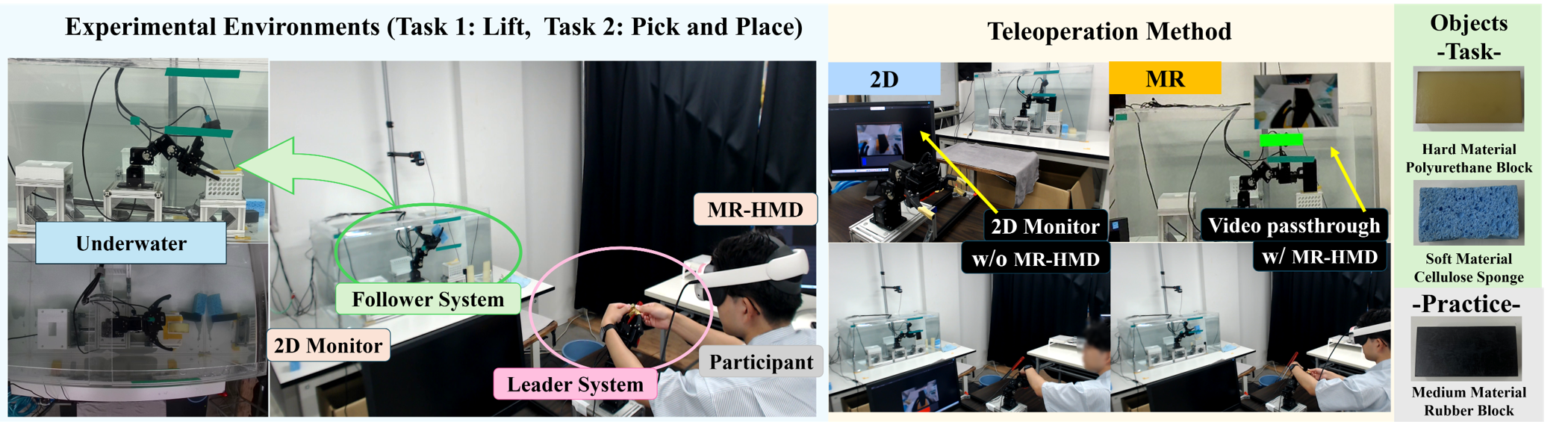}
    \caption{Experimental setup. Left: the operator sits beside the tank, so the
workspace is in front of them in both conditions --- seen directly under 2D Monitor and
through passthrough under MR-GLi. Centre: the two configurations, labelled
\emph{2D} and \emph{MR} in the photographs, i.e.\ 2D Monitor and MR-GLi. Right: the rigid polyurethane block
and compliant cellulose sponge used in trials, and the practice block.}
    \label{fig:setup}
\end{figure*}
\section{EXPERIMENTS}

\subsection{Study Design}\label{sec:study}
We used a within-subject design with display configuration as the independent variable.
The evaluation covered torque regulation, perceived gaze-shift burden, workload, and
usability. Perceived gaze-shift burden was the primary subjective outcome, and time
within the optimal torque range was the primary objective outcome.

\subsection{Experimental Setup and Display Conditions}\label{sec:conditions}
The two configurations deliver the same bilateral control, RTI encoding and
wrist-camera image, and differ only in how that information reaches the operator
(Fig.~\ref{fig:setup}, Table~\ref{tab:conditions}); the table's lower block lists the
factors that unavoidably covary with placement.

\begin{table}[t]
\centering
\caption{Comparison of the two display configurations.}
\label{tab:conditions}
\footnotesize
\renewcommand{\arraystretch}{1.12}
\setlength{\tabcolsep}{4pt}
\begin{tabular}{@{}lcc@{}}
\toprule
\textbf{Component} & \textbf{2D Monitor} & \textbf{MR-GLi} \\
\midrule
Four-channel bilateral control & \cmark & \cmark \\
RTI (identical encoding, thresholds) & \cmark & \cmark \\
Wrist-camera image of the gripper & \cmark & \cmark \\
\midrule
\multicolumn{3}{@{}l}{\emph{Covarying with placement:}} \\
RTI and camera registered to the gripper & -- & \cmark \\
Direct unmediated view of the tank & \cmark & passthrough \\
Gaze shift to a separate display & \cmark & -- \\
Head-borne weight, narrowed FOV & -- & \cmark \\
\bottomrule
\end{tabular}
\end{table}

In the \textbf{2D Monitor} condition the operator views the tank directly and reads both
elements --- the RTI and a wrist-camera close-up --- on a 24-inch monitor beside the
leader device. Because the two sit side by side, gripper state and torque can be read
together without looking at the tank, so this is a strong rather than a straw baseline;
the wrist camera covers neither the arm nor the approach path, so gross positioning is
still done from the direct view.

In the \textbf{MR-GLi} condition the operator wears a Quest~3 and views the same
workspace through color video passthrough, the two elements being the gripper-registered
panels, so both remain in view without a gaze shift. Viewing
geometry was fixed by the room layout but not instrumented.

\subsection{Participants and Procedure}
Twenty participants completed the study (9 female, 11 male; mean age 22.0 years, SD 2.7,
range 18--29). Five had never used an MR-HMD, twelve had used one a few times and three
frequently.
Written informed consent was obtained in accordance with institutional guidelines.

The follower arm was mounted at the bottom of a water tank; the leader stood on a desk
about 3\,m away in the same room, so head-motion parallax was available in both
conditions. All four joints were logged at approximately 1\,kHz, giving 160 trials
(20 participants $\times$ 2 conditions $\times$ 4 task--object combinations). 

\begin{figure}[t]
    \centering
    \includegraphics[width=0.9\linewidth]{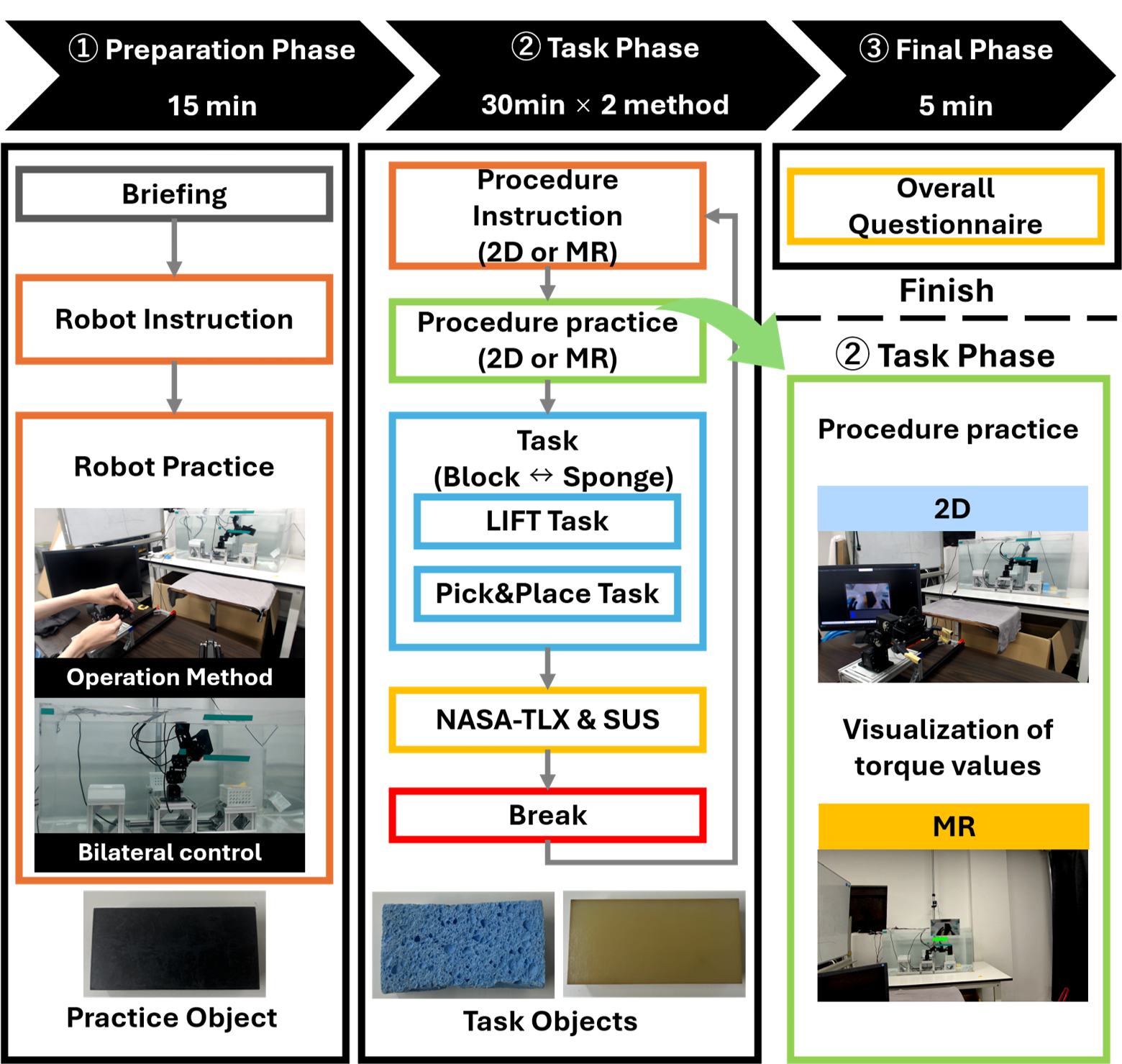}
\caption{Experimental protocol under the two display conditions.}
    \label{fig:protocol}
\end{figure}

\begin{figure}[t]
    \centering
    \includegraphics[width=\linewidth]{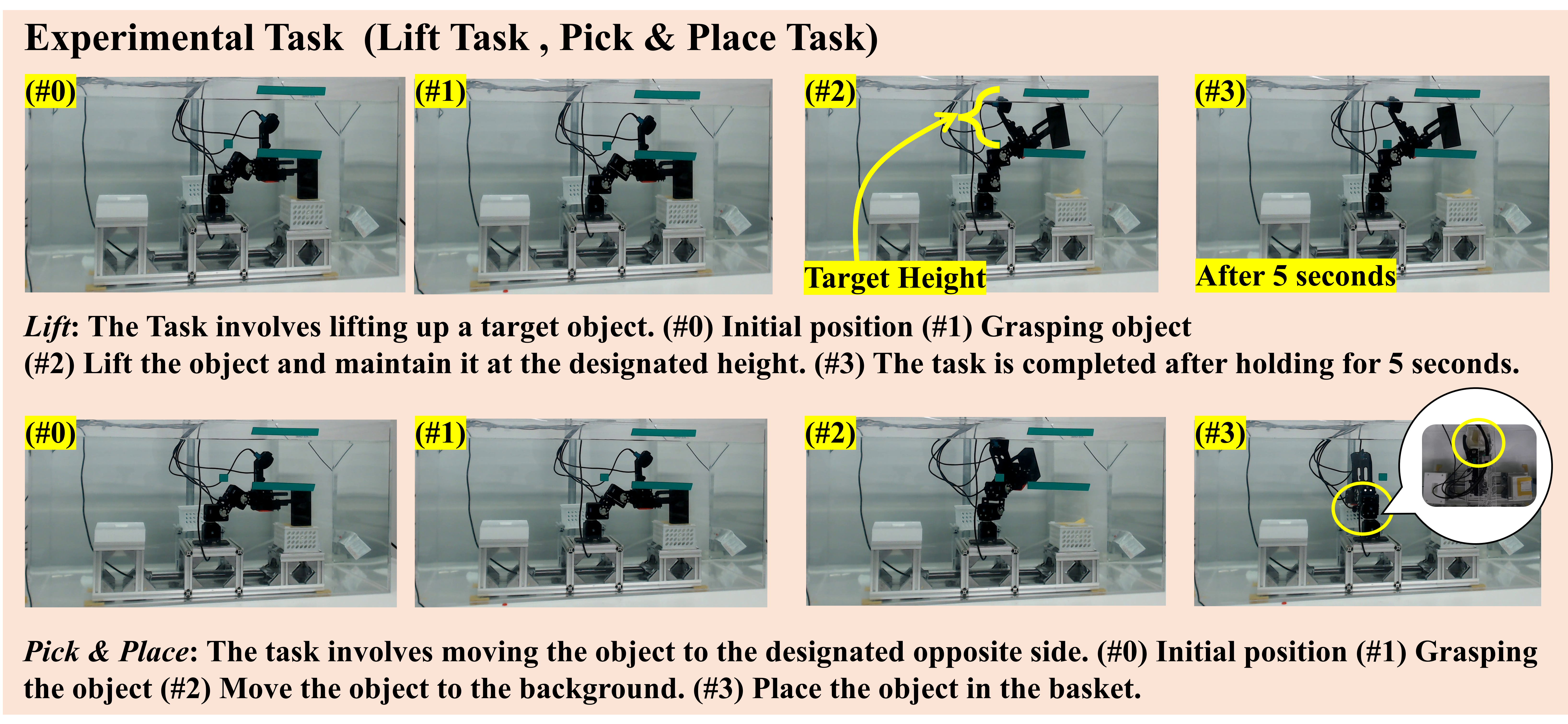}
    \caption{The two tasks. \emph{Lift}: grasp and hold the object at a marked
height for five seconds. \emph{Pick\&Place}: grasp, carry across the tank, release into
the basket. Both were performed on both objects.}
    \label{fig:task}
\end{figure}

The session followed the three-phase
protocol of~\cite{mrubi} (Fig.~\ref{fig:protocol}): preparation (15\,min), task
(30\,min per condition), and a questionnaire (5\,min). Practice used a rubber
block of a stiffness unlike either experimental object, and condition order was
counterbalanced, ten participants meeting each display first. Each condition comprised
\emph{Lift} and \emph{Pick\&Place} (Fig.~\ref{fig:task}) on a rigid polyurethane
block and a compliant cellulose sponge, chosen to span a wide stiffness range. The four task--object
combinations ran in a fixed sequence, identical for every participant and both
conditions.

\subsection{Evaluation Metrics}
Objective performance was evaluated using five torque-regulation measures and trial
time. Torque measures were computed over the task-specific evaluation interval:
\begin{itemize}
  \item \emph{Optimal} [\%]: percentage of time within the target torque range
  $[T_{\mathrm{low}},T_{\mathrm{high}}]$, designated as the primary objective outcome.
  \item \emph{Low} and \emph{High} [\%]: percentages of time below
  $T_{\mathrm{low}}$ and above $T_{\mathrm{high}}$, respectively.
  \item \emph{MAE} [N$\cdot$m]: mean absolute error relative to the target torque
  $T_{\mathrm{opt}}$.
  \item \emph{SD} [N$\cdot$m]: standard deviation of grasping torque over the
  evaluation interval.
  \item \emph{Trial time} [s]: elapsed time from grasp to release in
  \emph{Pick\&Place}. Timing was not evaluated for \emph{Lift}, which required a
  fixed five-second hold.
\end{itemize}

Subjective evaluation comprised the System Usability Scale (SUS)~\cite{sus},
NASA-TLX~\cite{nasa-tlx}, and a final comparative questionnaire. SUS and NASA-TLX were
administered after each display condition. SUS was scored using the standard procedure;
NASA-TLX included the overall workload score, weighted using pairwise subscale
comparisons, and the six individual subscale scores.

The final questionnaire assessed perceived ease of gaze shifting and task focus for
each configuration on five-point scales ranging from difficult to easy, and asked in a
forced choice which of the two configurations had been easier to use. Gaze-shift ease
was the primary subjective outcome. All items were collected retrospectively after
participants completed the two conditions.

Analyses were conducted at the participant level using
condition-wise averages across trials. Paired differences were
assessed using Wilcoxon signed-rank tests, with effect size
$r = |Z|/\sqrt{n}$. Tables report $p$-values and $r$.
Statistical significance was assessed at the .05 level.

\begin{table}[t]
\centering
\caption{Experimental results}
\label{tab:objective}
\small
\renewcommand{\arraystretch}{1.12}

\resizebox{\columnwidth}{!}{%
\begin{tabular}{lcccccc}
\toprule
 & \textbf{Optimal [\%]} & \textbf{Low [\%]} & \textbf{High [\%]} &
\textbf{MAE [N$\cdot$m]} & \textbf{SD [N$\cdot$m]} &
\textbf{Trial time [s]}$^{\ast}$ \\
\midrule
2D Monitor & 86.36 & 6.72 & 6.92 & 0.056 & 0.057 & 16.74 \\
MR-GLi     & 86.42 & 6.25 & 7.34 & 0.059 & 0.053 & 14.14 \\
\midrule
\multicolumn{7}{l}{\itshape Difference, MR-GLi $-$ 2D Monitor} \\
\quad $p$ & .546 & .498 & .475 & .409 & .409 & .053 \\
\quad $r$ & 0.14 & 0.16 & 0.17 & 0.19 & 0.19 & 0.43 \\
\bottomrule
\end{tabular}%
}
\end{table}

\subsection{Torque Regulation}\label{sec:torque}
Table~\ref{tab:objective} summarizes the objective results. No significant difference
was found between MR-GLi and the 2D Monitor in time within the optimal torque range
($p=.546$, $r=.14$), nor in any of the other torque-regulation measures
(all $p\geq.409$, $r\leq.19$).

Trial time was shorter with MR-GLi, but the difference was not statistically significant
($p=.053$, $r=.43$). Because trial time also depended on condition order
($p=.021$), this difference may partly reflect familiarization with the task.
No corresponding order effect was observed for torque-regulation accuracy
($p=.68$).

\begin{figure}[t]
\centering
\includegraphics[width=\linewidth]{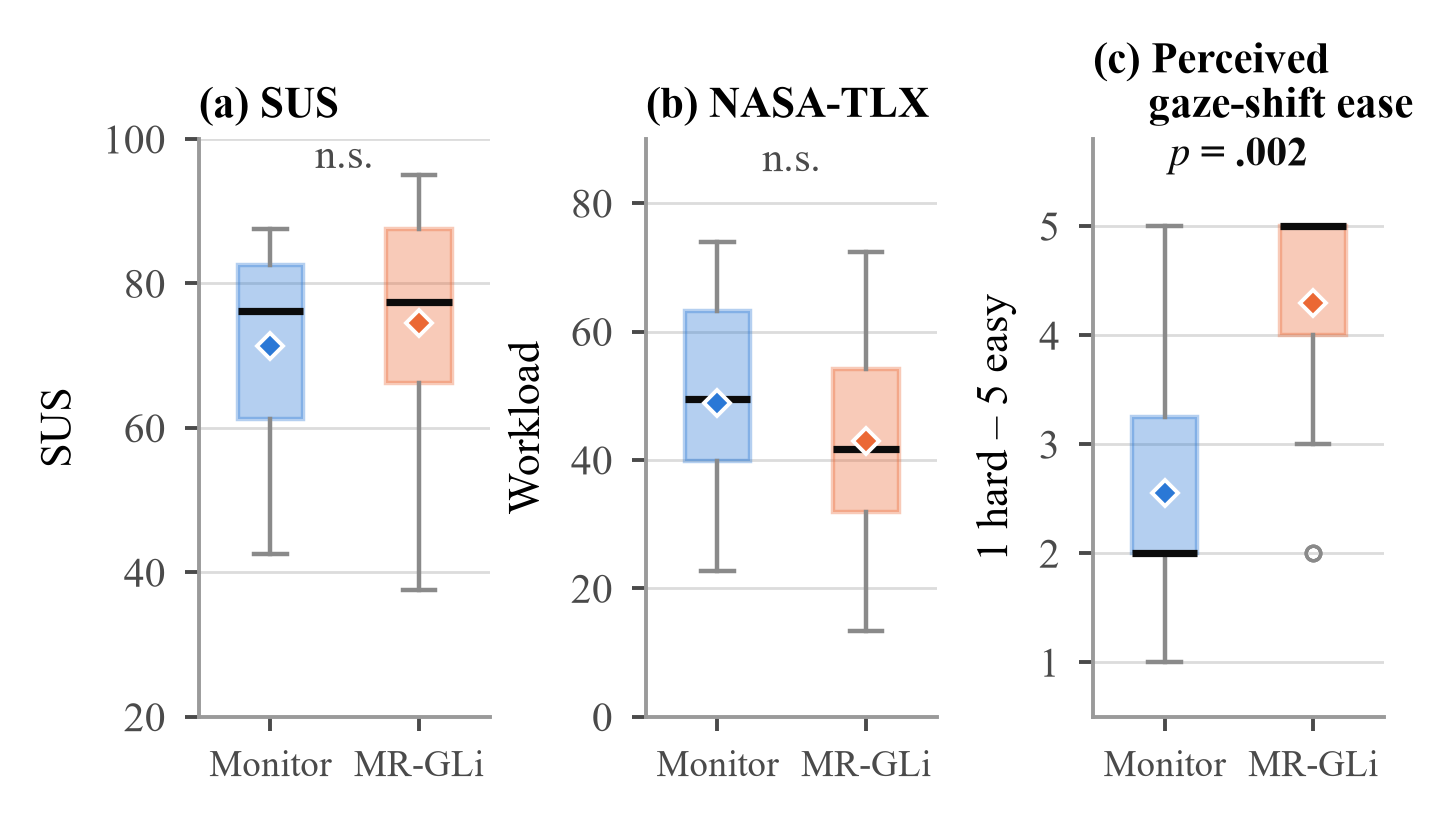}
\caption{Subjective measures by configuration ($n=20$).}
\label{fig:subjective}
\end{figure}

\begin{table}[t]
\centering
\caption{Subjective measures at participant level ($n=20$), median [IQR]. Higher is
better for SUS and the two five-point items, lower for NASA-TLX and its sub-scales.}
\label{tab:subjective}
\footnotesize
\renewcommand{\arraystretch}{1.15}
\setlength{\tabcolsep}{4.5pt}
\begin{tabular}{lcccc}
\toprule
\textbf{Measure} & \textbf{2D Monitor} & \textbf{MR-GLi} & \textbf{$r$} & \textbf{$p$} \\
\midrule
SUS                 & 76.2 [21.2] & 77.5 [21.2] & 0.15 & .506 \\
\addlinespace[3pt]
NASA-TLX            & 49.5 [23.4] & 41.7 [22.3] & 0.26 & .261 \\
\quad Mental        & 62.5 [41.2] & 37.5 [37.5] & 0.48 & .034 \\
\quad Physical      & 42.5 [32.5] & 40.0 [22.5] & 0.06 & .825 \\
\quad Temporal      & 22.5 [30.0] & 27.5 [26.2] & 0.04 & .888 \\
\quad Performance   & 32.5 [50.0] & 30.0 [60.0] & 0.00 & 1.000 \\
\quad Effort        & 45.0 [46.2] & 42.5 [36.2] & 0.22 & .346 \\
\quad Frustration   & 42.5 [50.0] & 35.0 [32.5] & 0.20 & .359 \\
\addlinespace[3pt]
Gaze-shift ease     & 2.0 [1.2]   & 5.0 [1.0]   & 0.68 & \textbf{.002} \\
Task focus          & 3.0 [2.0]   & 4.0 [2.0]   & 0.39 & .081 \\
\bottomrule
\end{tabular}
\end{table}

\subsection{Subjective Measures}
Fig.~\ref{fig:subjective} and Table~\ref{tab:subjective} summarize the subjective outcomes.
MR-GLi significantly improved perceived gaze-shift ease
($p = .002$, $r = .68$).

No significant differences were found in usability
(SUS: $p = .506$, $r = .15$) or overall workload
(NASA-TLX: $p = .261$, $r = .26$).
However, both measures numerically favored MR-GLi:
the median SUS score was slightly higher for MR-GLi
(77.5 vs.\ 76.2), while the median NASA-TLX score was lower
(41.7 vs.\ 49.5), indicating a tendency toward higher usability
and lower perceived workload.

Mental demand, a secondary outcome, also favored MR-GLi
($p = .034$, $r = .48$), with a lower median score
(37.5 vs.\ 62.5).
Task focus did not differ significantly
($p = .081$, $r = .39$), although the median rating was higher
for MR-GLi (4.0 vs.\ 3.0).
Overall, the strongest subjective benefit of MR-GLi was the
reduced perceived burden of gaze shifting, while the other
subjective measures showed no significant deterioration and
generally tended to favor MR-GLi.

\subsection{Discussion}
MR-GLi improved access to visual feedback while maintaining
torque-regulation performance comparable to the 2D Monitor.
Perceived gaze-shift ease significantly improved, suggesting
that the MR-GLi configuration can reduce the perceived burden
of shifting attention between the workspace and visual feedback.

Although overall usability and workload did not differ significantly, both SUS and
NASA-TLX numerically favored MR-GLi. Mental demand and task focus also tended to favor
MR-GLi, and fourteen of twenty participants found it easier to use. These results
suggest that gripper-linked MR overlays can improve access to visual feedback without
degrading torque-regulation performance, usability, or workload.

\section{CONCLUSIONS}
This paper compared MR-GLi with a conventional 2D monitor for underwater bilateral
teleoperation while keeping the controller and visual-feedback content consistent.
A counterbalanced within-subject study with twenty participants showed that MR-GLi
maintained torque-regulation performance comparable to the 2D monitor, while subjective
evaluation indicated reduced perceived burden in accessing visual feedback and comparable
usability and workload.

These results demonstrate the feasibility of gripper-linked MR overlays for underwater
bilateral teleoperation without degrading torque-regulation performance. Future work
will evaluate the approach with direct gaze measurements and under more challenging
conditions.

\end{document}